\documentclass[conference]{IEEEtran}
\IEEEoverridecommandlockouts
\usepackage{cite}
\usepackage{amsmath,amssymb,amsfonts}
\usepackage{algorithmic}
\usepackage{graphicx}
\usepackage{textcomp}
\usepackage{xcolor}
\def\BibTeX{{\rm B\kern-.05em{\sc i\kern-.025em b}\kern-.08em
    T\kern-.1667em\lower.7ex\hbox{E}\kern-.125emX}}

\usepackage{graphicx} 
\usepackage{multirow}
\usepackage{makecell}    
\usepackage{enumitem}
\usepackage{booktabs}

\usepackage{array, makecell, booktabs}
\usepackage{url}

\DeclareRobustCommand*{\IEEEauthorrefmark}[1]{%
    \raisebox{0pt}[0pt][0pt]{\textsuperscript{\footnotesize\ensuremath{#1}}}}
    
\begin{document}

\title{Fine-Grained Zero-Shot Composed Image Retrieval with Complementary Visual-Semantic Integration
}

\author{
    \IEEEauthorblockN{
        Yongcong Ye\IEEEauthorrefmark{1},
        Kai Zhang\IEEEauthorrefmark{1}\textsuperscript{*}\thanks{* Corresponding author.},
        Yanghai Zhang\IEEEauthorrefmark{1},
        Enhong Chen\IEEEauthorrefmark{1},
        Longfei Li\IEEEauthorrefmark{3},
        Jun Zhou\IEEEauthorrefmark{2}
    }
    \IEEEauthorblockA{\IEEEauthorrefmark{1} State Key Laboratory of Cognitive Intelligence, University of Science and Technology of China, Hefei, China}
    \IEEEauthorblockA{\IEEEauthorrefmark{2} Zhejiang University, Hangzhou, China}
    \IEEEauthorblockA{\IEEEauthorrefmark{3} Independent Researcher}
    \IEEEauthorblockA{\{ycongy, yhzhang0612\}@mail.ustc.edu.cn, \{kkzhang08, cheneh\}@ustc.edu.cn}
    \IEEEauthorblockA{lilongfei123@hotmail.com, junzhougucas@gmail.com}
}

\maketitle

\begin{abstract}
Zero-shot composed image retrieval (ZS-CIR) is a rapidly growing area with significant practical applications, allowing users to retrieve a target image by providing a reference image and a relative caption describing the desired modifications. Existing ZS-CIR methods often struggle to capture fine-grained changes and integrate visual and semantic information effectively. They primarily rely on either transforming the multimodal query into a single text using image-to-text models or employing large language models for target image description generation, approaches that often fail to capture complementary visual information and complete semantic context. To address these limitations, we propose a novel Fine-Grained Zero-Shot Composed Image Retrieval method with Complementary Visual-Semantic Integration (CVSI). Specifically, CVSI leverages three key components: (1) Visual Information Extraction, which not only extracts global image features but also uses a pre-trained mapping network to convert the image into a pseudo token, combining it with the modification text and the objects most likely to be added. (2) Semantic Information Extraction, which involves using a pre-trained captioning model to generate multiple captions for the reference image, followed by leveraging an LLM to generate the modified captions and the objects most likely to be added. (3) Complementary Information Retrieval, which integrates information extracted from both the query and database images to retrieve the target image, enabling the system to efficiently handle retrieval queries in a variety of situations. Extensive experiments on three public datasets (e.g., CIRR, CIRCO, and FashionIQ) demonstrate that CVSI significantly outperforms existing state-of-the-art methods. Our code is available at \url{https://github.com/yyc6631/CVSI}.
\end{abstract}


\section{Introduction}
Traditional image-text retrieval methods \cite{ref28,ref30,ref57} typically involve retrieving text from images or retrieving images from text, where the input is usually unimodal data. However, Composed Image Retrieval (CIR) \cite{ref3,ref4,ref5} focuses on more realistic scenarios. When a user wants to make specific modifications to an image and retrieve a target image that aligns with both the image content and the modification's semantic information, CIR is required. The input to CIR consists of multimodal data, including a reference image and modification text. It enables more precise retrieval based on the user's specific needs and has potential applications in fields such as E-commerce, among others \cite{ref43,zhang2021eatn,zhang2021multi}, which has led to growing attention.

\begin{figure}[!t]
  \centering
  \vspace{0.1cm}
  \includegraphics[width=\linewidth]{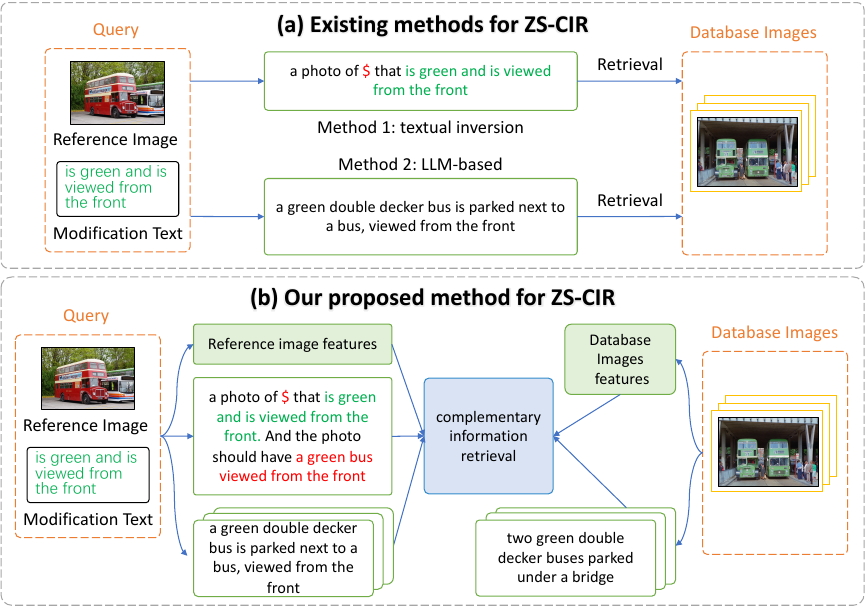}
  \vspace{-0.5cm}
  \caption{Comparison of methods for ZS-CIR. (a) Existing methods. (b) Our proposed method.}
  \label{figure 1}
  \vspace{-0.25cm}
\end{figure}

However, the CIR datasets typically requires training data in the form of triplets \( \langle I_r, T, I_t \rangle \), where \( I_r \) is the reference image, \( T \) is the modification text, and \( I_t \) is the target image. Unlike image-text paired datasets, collecting such triplets is challenging, which limits the widespread application of CIR in more domains. To address this data bottleneck, recent research has focused on Zero-Shot Composed Image Retrieval (ZS-CIR) task \cite{ref21}, which aims to generalize CIR models without relying on pre-collected triplet datasets. There are two main existing approaches to ZS-CIR, as shown in Figure \ref{figure 1}(a). The first method \cite{ref21,ref22,ref23,ref24,ref25,ref31} is based on textual inversion, which requires pre-training a mapping network. The pre-trained mapping network maps the image into a pseudo token, which is then concatenated with the modification text to form a single text for retrieval in the CLIP feature space. The second method \cite{ref26,ref27}, based on an LLM, bypasses pre-training by using a pre-trained captioning model to generate a caption for the reference image, which is then combined with the modification text to produce the target caption for retrieval. 

Although these methods have made significant progress in ZS-CIR, they overlook the complementary nature of visual and semantic information, focusing only on one aspect for complete extraction. As shown in Figure \ref{figure 1}(a), method 1 (i.e., textual inversion) combines the pseudo token and modification text into a single query using a fixed template for retrieval, which contains visual information but lacks complete and coherent semantic information (e.g., “a green double decker bus is parked next to a bus”). In contrast, method 2 (i.e., LLM-based) generates captions for the reference image, but struggles to fully capture the rich visual information (e.g., the shape and style of the double decker bus in the reference image). Moreover, both methods fail to enable fine-grained retrieval of the changed objects (e.g., retrieving “a green bus” instead of other objects in the image) and neglect the extraction and complementarity of both visual and semantic information in the target image. The ZS-CIR task is inherently a retrieval task that integrates both visual and semantic information. According to \cite{ref58}, simple modifications rely more on visual information, while complex ones depend mainly on semantic information for retrieval. Missing either type of information negatively impacts the model’s retrieval performance. Therefore, as shown in Figure \ref{figure 1}(b), we propose to integrate complementary visual and semantic information extracted separately from the Query and Database Images, which further enables fine-grained composed image retrieval.

\begin{figure*}[t]
  \centering
  \includegraphics[width=\linewidth]{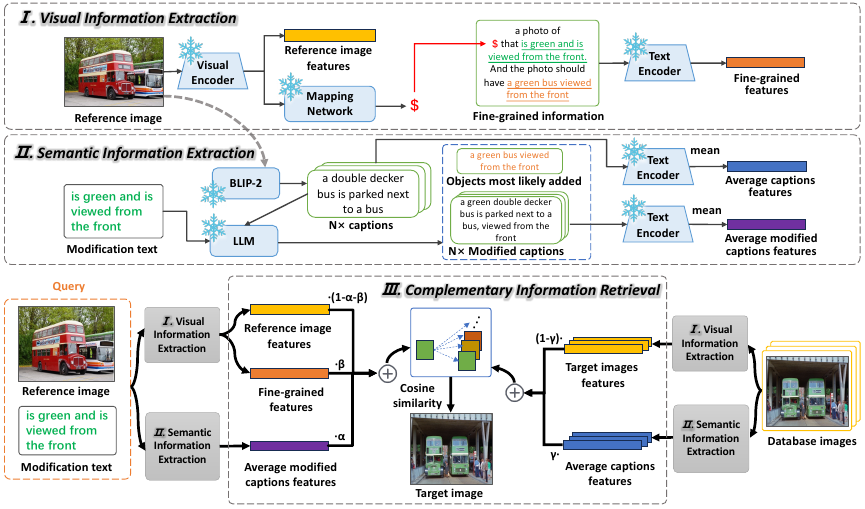}
  \caption{Overview of the proposed CVSI. The CVSI consists of three key components: (1) Visual Information Extraction, (2) Semantic Information Extraction, and (3) Complementary Information Retrieval.}
  \label{figure 2}
\end{figure*}

Indeed, there are many technical challenges inherent in designing effective solutions to integrate visual-semantic information into ZS-CIR process. \textbf{\emph{1) How to extract rich and complete visual and semantic information?}} In previous ZS-CIR methods, while the use of pseudo tokens allows for relatively complete extraction of visual information, some loss of visual information still occurs during the mapping process. For LLM-based methods, Yang et al. \cite{ref27} proposed a method for generating dense captions in the ZS-CIR task, which captures rich semantic information. \textbf{\emph{2) How to achieve fine-grained retrieval, enabling retrieval at the object level?}} Most prior methods \cite{ref21,ref22,ref24,ref25} rely on global features for retrieval, which miss fine-grained details and negatively impact retrieval performance. Some methods \cite{ref23,ref31} extract local features by training a model for retrieval, but this increases both model complexity and training time. \textbf{\emph{3) How to effectively integrate the visual and semantic information from both the Query and Database Images for retrieval?}} Previous approaches focus on extracting either visual or semantic information from the Query, but integrating multiple sources of visual and semantic information for retrieval presents a significant challenge.

To address these challenges, we propose a novel fine-grained zero-shot composed image retrieval method with Complementary Visual-Semantic Integration (CVSI). As shown in Figure \ref{figure 2}, our CVSI consists of three key components: (1) Visual Information Extraction, (2) Semantic Information Extraction, and (3) Complementary Information Retrieval. \textbf{\emph{For Challenge 1}}, to extract comprehensive visual information, we leverage the mapping network in LinCIR \cite{ref25}, a text-based training method, to generate pseudo tokens from images. Additionally, CLIP is employed to extract global image features as supplementary visual information. To capture complete semantic information, we use BLIP-2 \cite{ref32} to generate dense captions, as described in \cite{ref27}. Then, an LLM is employed to generate modified captions, enriching the semantic information. \textbf{\emph{For Challenge 2}}, we use an LLM to generate the objects most likely to be added based on the modification text and captions, which allows us to pinpoint changes at the object level rather than the entire image, enabling fine-grained retrieval. \textbf{\emph{For Challenge 3}}, in the real-world ZS-CIR task, the combinations of images and texts are often complex and diverse, and the impact of visual and semantic information on the final retrieval results varies. Therefore, we assign weights based on the nature of the query and separately perform weighted summation on the query and target features, followed by retrieval in the CLIP feature space. 

Our main contributions can be summarized as follows:
\begin{itemize}
  \item We propose a novel fine-grained zero-shot composed image retrieval method with Complementary Visual-Semantic Integration (CVSI). It not only fully considers the complementarity of visual and semantic information in the query but also in the database images, enabling comprehensive retrieval.
  \item We propose a fine-grained retrieval method that uses an LLM to generate the objects most likely to be added in the target image and incorporates them as part of the query features, which enables simple yet efficient fine-grained retrieval, enhancing retrieval performance.
  \item We propose a complementary information fusion and retrieval method that assigns weights based on the nature of the query, fully considering the varying importance of visual and semantic information in different contexts, which enables better adaptation to the real-world ZS-CIR task.
  \item Extensive experiments on three public datasets validate the effectiveness of our proposed method, with CVSI achieving new state-of-the-art results on most metrics.
\end{itemize}

\section{Related work}

\subsection{Composed Image Retrieval}

The field of image retrieval has garnered significant attention from many researchers \cite{ref1,ref2}. One area that has remarkable progress recently is Composed Image Retrieval (CIR) \cite{ref3,ref4,ref5}, a task focused on retrieving target images based on queries composed of a reference image and a modification text. Existing CIR methods can be primarily divided into two categories. The first category \cite{ref4,ref5,ref45,ref46} adopts standard deep learning methods to extract features from both visual and textual modalities. On the other hand, the second category \cite{ref49,ref50,ref58} employs vision-language pre-trained models for extracting features, such as CLIP \cite{ref28}. However, these methods rely on supervised pre-training, which requires a large amount of manually labeled triplets for training, making it time-consuming and labor-intensive.

Thus, Zero-Shot CIR \cite{ref21} is introduced because it does not require manually annotated triplets for training. Methods for Zero-Shot CIR can be broadly categorized into two types. One approach \cite{ref21,ref22,ref23,ref24,ref25,ref31} involves using a pre-trained mapping model to transform the reference image into a pseudo token, which is then combined with the modification text for retrieval. Specifically, Gu et al. \cite{ref25} proposed a method that trains the mapping model using only text. The other approach \cite{ref26,ref27} involves captioning the reference image using a pre-trained Vision-Language Model (VLM), and then using an LLM to recombine the caption based on the modification text for subsequent retrieval.

However, the approach using mapping models often results in retrieval queries that lack sufficient contextual semantic coherence. When the modification text is complex, retrieval performance tends to degrade. On the other hand, another approach of converting reference image information into captions often struggles to capture the full range of information in an image, or to highlight its key aspects. This results in a lack of refinement in visual information, which can lead to poor retrieval performance when the reference image itself is crucial. To overcome this challenge, we propose a method that complements visual and semantic information, incorporating changed details to enhance fine-grained retrieval.

\subsection{Vision-Language Pretrained Models for CIR}
Pretrained models have shown remarkable effectiveness across domains \cite{ref101,zhang2022incorporating}. Recently, Vision-Language Pretrained (VLP) models trained on large-scale image-text pairs have gained attention for their versatility and strong performance on various downstream tasks \cite{ref28,ref30,ref100}. Among these models, CLIP \cite{ref28} stands out as one of the most representative, demonstrating excellent performance across a wide range of retrieval tasks, including CIR. Additionally, BLIP \cite{ref29}, primarily designed for multimodal understanding and generation tasks, has also shown impressive cross-modal alignment and generation capabilities across multiple tasks. In this paper, for the Zero-Shot CIR task, we leverage the BLIP-2 \cite{ref32} model to generate captions for reference images and use CLIP to extract features for retrieval, fully exploiting the strengths of VLP models.

\section{Method}
In this section, we first formulate the problem and then provide a detailed description of our proposed CVSI framework. As shown in Figure \ref{figure 2}, CVSI consists of three main components: Visual Information Extraction, Semantic Information Extraction and Complementary Information Retrieval.
\subsection{Problem Formulation}
ZS-CIR is essentially a multimodal retrieval task. Unlike traditional multimodal retrieval tasks, its input includes both image and text modalities. Given a reference image \( I_r \) and a modification text \( T \), the goal is to retrieve a target image \( I_t \) from an image database \( \mathcal{D} = \{ I_k \}_{k=1}^{K} \), where the target image is required to be visually and semantically consistent with the reference image \( I_r \) modified by \( T \). Since this is a zero-shot task, it requires that no training samples are available.

\subsection{Visual Information Extraction}
In this component, we utilize a pre-trained VLM, such as CLIP or BLIP, which includes a vision encoder and a text encoder as feature extractors. For the reference image \( I_r \), the frozen visual encoder \( f_v \) extracts the global features of the image as \( q_v = f_v(I_r) \in \mathbb{R}^{d \times 1} \), where \( d \) is the feature dimension. Then, following LinCIR \cite{ref25}, we leverage a mapping network trained solely on text, which maps the image to a pseudo token, denoted as
\begin{equation}
\$ = \Phi(q_v), 
\end{equation}
where \( \$ \) is the pseudo token and \( \Phi \) is the frozen mapping network. Then, we concatenate the pseudo token with the modification text \( T \) and the objects most likely added to form a prompt in the following format: "a photo of \$ that [\( T \)]. And the photo should have [objects most likely added]", which serves as the fine-grained information \( F_G \). Finally, we extract the text features from the \( F_G \) by passing it through the frozen text encoder \( f_t \), resulting in
\begin{equation}
q_f = f_t(F_G), 
\end{equation}
where \( q_f \) represents the fine-grained features. Similarly, for each image \( I_k \) in the database \( \mathcal{D} \), we extract the visual features \( t_v = f_v(I_k) \in \mathbb{R}^{d \times 1} \) for subsequent retrieval.

\subsection{Semantic Information Extraction}
In this component, we aim to employ an image captioning model to derive a textual description for the reference image or the database images. Specifically, we utilize a pre-trained captioning model to generate natural language captions for the images, and for this, we have chosen BLIP-2 \cite{ref32}. Furthermore, to obtain dense captions, we follow LDRE \cite{ref27} and employ nucleus sampling \cite{ref33} during the caption generation process to enhance the diversity and semantic richness of the generated captions. The generation process of multiple captions can be described as follows:
\begin{equation}
\mathcal{C}^B = \{ \mathcal{C}_i^B \}_{i=1}^{N} = \text{BLIP-2}(I_r),
\end{equation}
where \( \mathcal{C}^B \) is the set of captions generated from the reference image. Similarly, the following equation holds:
\begin{equation}
\mathcal{C}^T = \{ \mathcal{C}_i^T \}_{i=1}^{N} = \text{BLIP-2}(I_k),
\end{equation}
where \( \mathcal{C}^T \) is the set of captions generated from \( I_k \) in the database. For the query, by generating multiple captions for the reference image, we can capture rich semantic information from the reference image, thus avoiding the potential loss of key semantic information in the query when only one caption is used. For the database images, the rich semantic information from the images to be retrieved is also involved in the final retrieval process. Unlike previous methods where only image features are used in the final retrieval, we pass the set of captions \( \mathcal{C}^T \) through the frozen text encoder \( f_t \), then average all the text features to obtain the average caption features \( t_c \), which helps to obtain a more balanced semantic representation. Formally, we have:
\begin{equation}
t_c = \frac{1}{N} \sum_{i=1}^{N} f_t(\mathcal{C}_i^T).
\end{equation}

When extracting semantic information, for the query, if we only perform feature extraction on the modification text, the semantic information will solely reflect the modified parts, thereby losing the relatively unchanged parts. On the other hand, if we only extract features from the reference image captions, the most crucial modification information may be lost. Although some previous methods \cite{ref21,ref22,ref23,ref24,ref25,ref31} have used fixed templates to combine them, such approaches result in rigid sentence structures and lack semantic coherence. Therefore, following previous LLM-based methods, we also use an LLM to generate multiple modified captions for each query. Additionally, if we can identify which objects are most likely to be added to the reference image in order to approximate the target image, we can direct the retrieval towards specific objects, making the retrieval more intuitive and enabling fine-grained search capabilities.

\begin{figure}[h]
  \centering
  \includegraphics[width=\linewidth]{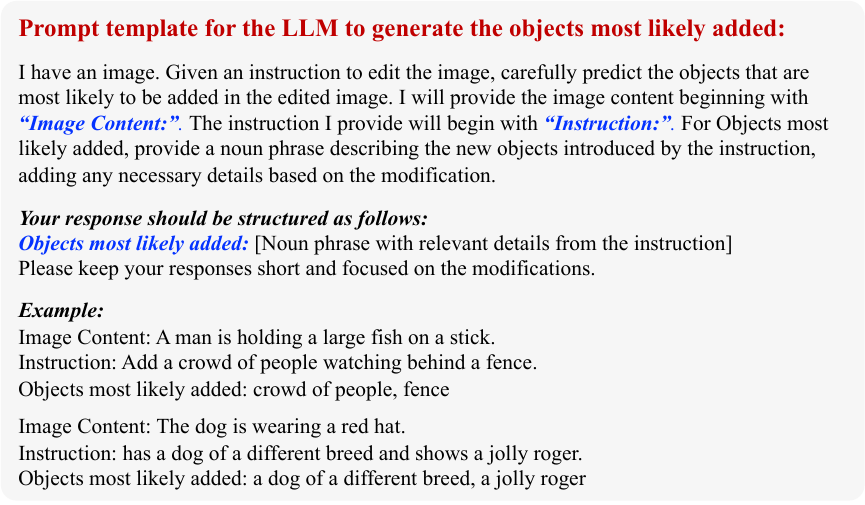}
  \caption{The description of the designed prompt template \( p_a \).}
  \label{figure 3}
\end{figure}

For the objects most likely added, we define a simple prompt template \( p_a \) that combines the first reference image caption \( \mathcal{C}_1^B \) with the modification text \( T \), forming a complete prompt. This prompt is then used by the LLM to generate the objects most likely added. Formally, we have:
\begin{equation}
F_{add} = \text{LLM}(p_a(\mathcal{C}_1^B, T)),
\end{equation}
where \( F_{add} \) refers to the objects most likely added. Additionally, the description of the prompt template \( p_a \) is shown in Figure \ref{figure 3}.

For generating modified captions, we adopt the prompt template \( p_m \) as used in \cite{ref26}, which can be used to combine \( \mathcal{C}_i^B \) (where \( i = 1, \dots, N \)) with the modification text \( T \) to form a complete prompt. This prompt is then used by the LLM to generate a set \( \mathcal{C}^M \) of \( N \) modified captions. We then pass \( \mathcal{C}^M \) through the frozen text encoder \( f_t \), and average all the text features to obtain the average modified caption features \( q_m \), ensuring that the final semantic features used in retrieval are balanced across multiple key pieces of information.
Thus, we have:
\begin{equation}
\left\{
\begin{array}{l}
  \mathcal{C}_i^M = \text{LLM}(p_m(\mathcal{C}_i^B, T)), \quad \text{for} \quad i = 1, \dots, N \\[10pt]
  q_m = \frac{1}{N} \sum_{i=1}^{N} f_t(\mathcal{C}_i^M),
\end{array}
\right.
\end{equation}
where \( \mathcal{C}_i^M \) represents the \( i \)-th modified caption, and \( q_m \) represents the final average modified caption features involved in retrieval.

\subsection{Complementary Information Retrieval}
Whether it is the previous textual inversion method or the LLM-based method, in the end, image retrieval is performed through text, which essentially transforms the ZS-CIR task into a unimodal retrieval task, thus mitigating the issues caused by misalignment between modalities. However, these methods overlook the fact that ZS-CIR is inherently a multimodal retrieval task. If more modalities are involved in the final retrieval process, and the retrieved target can provide more information (e.g., shifting from providing only visual information to offering both visual and semantic information), retrieval accuracy can be further improved. Moreover, in the real-world CIR task, there are various scenarios. In cases with complex modifications, many visual properties of the reference image need to be altered, making the modification text crucial. In contrast, in cases with simple modifications, only minor aspects of the reference image need to be changed, thus making the reference image play a more dominant role. To address these diverse situations, different modalities should be assigned different weights based on actual needs during the retrieval process.

In this component, we use the input reference image and modification text as the query. The query is passed into the visual information extraction component to obtain the reference image features \( q_v \) and fine-grained features \( q_f \). And it is also passed into the semantic information extraction component to obtain the average modified captions features \( q_m \). We then assign weights to these features and compute a weighted sum, which results in a feature that participates in the retrieval process. Formally, we have:
\begin{equation}
q = \alpha \cdot q_m + \beta \cdot q_f + (1 - \alpha - \beta) \cdot q_v,
\end{equation}
where \( q \) represents the final query features involved in retrieval, and \( \alpha \) and \( \beta \) are the weights of the corresponding features.

Similarly, for each image \( I_k \) in the Database (where \( k = 1, \dots, K \), and \( K \) is the number of images in the database), we input it into the visual information extraction component to obtain the target image features \( t_v \), and into the semantic information extraction component to obtain the average captions features \( t_c \). We then assign weights to these features and perform a weighted sum, ultimately obtaining the feature representation of each image in the database for retrieval. Therefore, we have:
\begin{equation}
t = \gamma \cdot t_c + (1 - \gamma) \cdot t_v,
\end{equation}
where \( t \) represents the final feature representation of each image in the database, and \( \gamma \) is the weight of the corresponding feature. Finally, for each image \( I_k \) in the database, a feature representation \( t_k \) is obtained. During retrieval, we compute the cosine similarity between \( q \) and each \( t_k \) in the CLIP feature space, and select the image with the highest similarity as the target image:

\begin{equation}
    I_t = \arg \max_{I_k} \frac{\mathbf{q} \cdot \mathbf{t}_k}{\|\mathbf{q}\| \|\mathbf{t}_k\|},
\end{equation}
where \( I_t \) is the target image.

\section{Experiment}
\begin{table*}[t]
  \centering
  \caption{Performance comparison on CIRCO and CIRR test sets. The best-performing results are shown in bold, and the second-best results are highlighted with an underline.}
    \begin{tabular}{ccll|cccc|ccc|ccc}
    \toprule
    \multicolumn{2}{c}{\multirow{3}[4]{*}{Backbone}} & \multicolumn{2}{c|}{\multirow{3}[4]{*}{Method}} & \multicolumn{4}{c|}{CIRCO}    & \multicolumn{6}{c}{CIRR} \\
\cmidrule{5-14}    \multicolumn{2}{c}{} & \multicolumn{2}{c|}{} & \multicolumn{4}{c|}{mAP@k}    & \multicolumn{3}{c|}{Recall@k} & \multicolumn{3}{c}{Rs@k} \\
    \multicolumn{2}{c}{} & \multicolumn{2}{c|}{} & k=5   & k=10  & k=25  & k=50  & k=1   & k=5   & k=10  & k=1   & k=2   & k=3 \\
    \midrule
    \multicolumn{2}{c}{\multirow{6}[2]{*}{ViT-B/32}} & \multicolumn{2}{l|}{PALAVRA{\footnotesize (ECCV'22)}} & 4.61  & 5.32  & 6.33  & 6.80  & 16.62  & 43.49  & 58.51  & 41.61  & 65.30  & 80.94  \\
    \multicolumn{2}{c}{} & \multicolumn{2}{l|}{SEARLE{\footnotesize (ICCV'23)}} & 9.35  & 9.94  & 11.13  & 11.84  & 24.00  & 53.42  & 66.82  & 54.89  & 76.60  & 88.19  \\
    \multicolumn{2}{c}{} & \multicolumn{2}{l|}{SEARLE-OTI{\footnotesize (ICCV'23)}} & 7.14  & 7.83  & 8.99  & 9.60  & 24.27  & 53.25  & 66.10  & 54.10  & 75.81  & 87.33  \\
    \multicolumn{2}{c}{} & \multicolumn{2}{l|}{CIReVL{\footnotesize (ICLR'24)}} & 14.94  & 15.42  & 17.00  & 17.82  & 23.94  & 52.51  & 66.00  & 60.17  & 80.05  & 90.19  \\
    \multicolumn{2}{c}{} & \multicolumn{2}{l|}{LDRE{\footnotesize (SIGIR'24)}} & \underline{17.96}  & \underline{18.32}  & \underline{20.21}  & \underline{21.11}  & \underline{25.69}  & \underline{55.13}  & \underline{69.04}  & \underline{60.53}  & \underline{80.65}  & \underline{90.70}  \\
    \multicolumn{2}{c}{} & \multicolumn{2}{l|}{CVSI{\footnotesize (Ours)}} & \textbf{21.67}\hspace{0pt}  & \textbf{22.47}\hspace{0pt}  & \textbf{24.46}\hspace{0pt}  & \textbf{25.44}\hspace{0pt}  & \textbf{31.64}\hspace{0pt}  & \textbf{62.43}\hspace{0pt}  & \textbf{74.19}\hspace{0pt}  & \textbf{64.17}\hspace{0pt}  & \textbf{82.48}\hspace{0pt}  & \textbf{91.69}\hspace{0pt}  \\
    \midrule
    \multicolumn{2}{c}{\multirow{10}[2]{*}{ViT-L/14}} & \multicolumn{2}{l|}{Image-only} & 1.28  & 1.70  & 2.35  & 2.69  & 3.64  & 12.75  & 23.32  & 11.58  & 31.41  & 45.26  \\
    \multicolumn{2}{c}{} & \multicolumn{2}{l|}{Text-only} & 2.63  & 2.85  & 3.30  & 3.58  & 20.51  & 43.21  & 55.08  & 60.39  & 80.02  & \underline{90.05}  \\
    \multicolumn{2}{c}{} & \multicolumn{2}{l|}{Captioning} & 1.65  & 1.96  & 2.42  & 2.71  & 4.05  & 15.88  & 25.69  & 20.87  & 40.60  & 60.89  \\
    \multicolumn{2}{c}{} & \multicolumn{2}{l|}{Pic2Word{\footnotesize (CVPR'23)}} & 8.72  & 9.51  & 10.64  & 11.29  & 23.90  & 51.70  & 65.30  & 53.76  & 74.46  & 87.08  \\
    \multicolumn{2}{c}{} & \multicolumn{2}{l|}{SEARLE{\footnotesize (ICCV'23)}} & 11.68  & 12.73  & 14.33  & 15.12  & 24.24  & 52.48  & 66.29  & 53.76  & 75.01  & 88.19  \\
    \multicolumn{2}{c}{} & \multicolumn{2}{l|}{SEARLE-OTI{\footnotesize (ICCV'23)}} & 10.18  & 11.03  & 12.72  & 13.67  & 24.87  & 52.31  & 66.29  & 53.80  & 74.31  & 86.94  \\
    \multicolumn{2}{c}{} & \multicolumn{2}{l|}{LinCIR{\footnotesize (CVPR'24)}} & 12.59  & 13.58  & 15.00  & 15.85  & 25.04  & 53.25  & 66.68  & 57.11  & 77.37  & 88.89  \\
    \multicolumn{2}{c}{} & \multicolumn{2}{l|}{CIReVL{\footnotesize (ICLR'24)}} & 18.57  & 19.01  & 20.89  & 21.80  & 24.55  & 52.31  & 64.92  & 59.54  & 79.88  & 89.69  \\
    \multicolumn{2}{c}{} & \multicolumn{2}{l|}{LDRE{\footnotesize (SIGIR'24)}} & \underline{23.35}  & \underline{24.03}  & \underline{26.44}  & \underline{27.50}  & \underline{26.53}  & \underline{55.57}  & \underline{67.54}  & \underline{60.43}  & \underline{80.31}  & 89.90  \\
    \multicolumn{2}{c}{} & \multicolumn{2}{l|}{CVSI{\footnotesize (Ours)}} & \textbf{26.89}\hspace{0pt}  & \textbf{28.24}\hspace{0pt}  & \textbf{30.79}\hspace{0pt}  & \textbf{31.89}\hspace{0pt}  & \textbf{33.81}\hspace{0pt}  & \textbf{63.64}\hspace{0pt}  & \textbf{75.04}\hspace{0pt}  & \textbf{65.81}\hspace{0pt}  & \textbf{82.92}\hspace{0pt}  & \textbf{92.17}\hspace{0pt}  \\
    \midrule
    \multicolumn{2}{c}{\multirow{6}[2]{*}{ViT-G/14}} & \multicolumn{2}{l|}{Pic2Word{\footnotesize (CVPR'23)}} & 5.54  & 5.59  & 6.68  & 7.12  & 30.41  & 58.12  & 69.23  & \underline{68.92}  & 85.45  & 93.04  \\
    \multicolumn{2}{c}{} & \multicolumn{2}{l|}{SEARLE{\footnotesize (ICCV'23)}} & 13.20  & 13.85  & 15.32  & 16.04  & 34.80  & 64.07  & 75.11  & 68.72  & 84.70  & 93.23  \\
    \multicolumn{2}{c}{} & \multicolumn{2}{l|}{LinCIR{\footnotesize (CVPR'24)}} & 19.71  & 21.01  & 23.13  & 24.18  & 35.25  & 64.72  & 76.05  & 63.35  & 82.22  & 91.98  \\
    \multicolumn{2}{c}{} & \multicolumn{2}{l|}{CIReVL{\footnotesize (ICLR'24)}} & 26.77  & 27.59  & 29.96  & 31.03  & 34.65  & 64.29  & 75.06  & 67.95  & 84.87  & 93.21  \\
    \multicolumn{2}{c}{} & \multicolumn{2}{l|}{LDRE{\footnotesize (SIGIR'24)}} & \underline{31.12}  & \underline{32.24}  & \underline{34.95}  & \underline{36.03}  & \underline{36.15}  & \underline{66.39}  & \underline{77.25}  & 68.82  & \underline{85.66}  & \underline{93.76}  \\
    \multicolumn{2}{c}{} & \multicolumn{2}{l|}{CVSI{\footnotesize (Ours)}} & \textbf{33.51}\hspace{0pt}  & \textbf{34.96}\hspace{0pt}  & \textbf{37.86}\hspace{0pt}  & \textbf{38.98}\hspace{0pt}  & \textbf{41.37}\hspace{0pt}  & \textbf{72.58}\hspace{0pt}  & \textbf{82.39}\hspace{0pt}  & \textbf{71.35}\hspace{0pt}  & \textbf{87.54}\hspace{0pt}  & \textbf{94.68}\hspace{0pt}  \\
    \bottomrule
    \end{tabular}%
  \label{tab:CIRCO and CIRR}%
\end{table*}%

\subsection{Experimental Settings}
\subsubsection{Evaluation Datasets}
To evaluate the performance of our CVSI in various downstream CIR tasks, following \cite{ref27}, we conduct our experiments on three public datasets:
\begin{itemize}[left=0pt]
    \item \textbf{CIRR}\cite{ref34} consists of 36,554 queries, each paired with a target image. Our model was evaluated on the test set of CIRR, which contains 4.1K triplets. 
    \item \textbf{CIRCO}\cite{ref22} is an open-domain dataset derived from the COCO dataset \cite{ref37}. Unlike CIRR, each sample in CIRCO consists of a reference image, a modification text, and multiple target images. And each query has an average of 4.53 ground truths. We evaluated our model on the test set, which comprises 800 samples. 
    \item \textbf{FashionIQ}\cite{ref35} is a fashion-focused dataset, which includes 77,683 images,  organized into three subcategories: Dress, Shirt, and Toptee. We assessed our model using the validation set, which contains a total of 6K triplets across the three categories.
\end{itemize}

\subsubsection{Implementation Details}
For the mapping network, we adopt the approach used in LinCIR \cite{ref25}. For captioning, we follow \cite{ref27} and utilize the pre-trained BLIP-2 \cite{ref32} with the OPT-6.7b large language model. The number of captions \(N\) is set to 15, as in \cite{ref27}. For the LLM, we use GPT-3.5-Turbo \cite{ref38} following prior work \cite{ref26, ref27}, and additionally perform scalability experiments with LLama2-70B \cite{ref41} and GPT-4o. For the VLM, we ensure fair comparison with prior methods by using CLIP models from OpenAI (ViT-B/32, ViT-L/14) and OpenCLIP (ViT-G/14) \cite{ref39}. Regarding hyperparameters, for datasets like CIRCO and CIRR, which involve complex modification texts, we set \( \alpha = 0.6 \), \( \beta = 0.4 \), and \( \gamma = 0.2 \). For FashionIQ, which relies more on visual information, we set \( \alpha = 0.2 \), \( \beta = 0.6 \), and \( \gamma = 0.1 \). The model was implemented in PyTorch using one NVIDIA V100 GPU.

\subsubsection{Evaluation Metrics}
We followed the standard evaluation protocols for each dataset. For CIRCO, we chose mean Average Precision (mAP) as a more precise metric, specifically mAP@k, where \( k \in \{5, 10, 25, 50\} \) and k denotes the number of top-ranked retrieval results being evaluated. For CIRR, we employed Recall@k (where \( k \in \{1, 5, 10\} \)) and Recall\(_{\text{subset}}\)@k (where \( k \in \{1, 2, 3\} \)) as evaluation metrics. Notably, Recall\(_{\text{subset}}\)@k limits the candidate target images to those semantically similar to the correct target image, helping to mitigate the issue of false negatives. For FashionIQ, we used Recall at rank \( K \) (Recall@K) as the evaluation metric, specifically adopting R@10 and R@50. Additionally, we computed the average performance across the three subsets of different categories to assess overall performance.

\subsubsection{Baselines}
To thoroughly evaluate the effectiveness of our method, we have adopted the following baselines and briefly summarized their key concepts. All of these methods ultimately perform retrieval in the CLIP feature space.
\begin{itemize}[left=0pt]
    \item \textbf{Image-only}. It directly uses the visual features of the reference image to retrieve the most similar target image.
    \item \textbf{Text-only}. It directly uses the textual features of the modification text to retrieve the most similar target image. 
    \item \textbf{Captioning}. The caption generated by BLIP-2 for the reference image is concatenated with the modification text to form a single query for retrieval.
    \item \textbf{PALAVRA} \cite{ref40}. It involves performing textual inversion through a mapping function, followed by optimizing the pseudo token.
    \item \textbf{Pic2Word} \cite{ref21}. It employs a textual inversion network to map the reference image into a pseudo token, which is then combined with the modification text to form a single query for retrieval.  
    \item \textbf{SEARLE / SEARLE-OTI} \cite{ref22}. The SEARLE-OTI framework employs an optimized textual inversion approach to learn a pseudo token. SEARLE improves upon this by learning a more efficient mapping network, distilling knowledge from SEARLE-OTI.
    \item \textbf{LinCIR} \cite{ref25}. It trains a textual inversion network solely through text to map the reference image into a pseudo token, which is then combined with the modification text to form a single query for retrieval.
    \item \textbf{CIReVL} \cite{ref26}. It uses a captioning model to generate a caption for each reference image, then employs an LLM to generate a modified caption for retrieval.
    \item \textbf{LDRE} \cite{ref27}. It uses a captioning model to generate multiple captions for each reference image, then employs an LLM to generate multiple modified captions, which are weighted and summed for retrieval.
\end{itemize}

\begin{table*}[t]
  \centering
  \caption{Performance comparison on FashionIQ validation set. The best-performing results are shown in bold, and the second-best results are highlighted with an underline.}
    \begin{tabular}{cclcccccccc}
    \toprule
    \multicolumn{2}{c}{\multirow{2}[4]{*}{Backbone}} & \multicolumn{1}{c}{\multirow{2}[4]{*}{Method}} & \multicolumn{2}{c}{Shirt} & \multicolumn{2}{c}{Dress} & \multicolumn{2}{c}{Topee} & \multicolumn{2}{c}{Average} \\
    \cmidrule{4-11}    \multicolumn{2}{c}{} &       & R@10 & R@50 & R@10 & R@50 & R@10 & R@50 & R@10 & R@50 \\
    \midrule
    \multicolumn{2}{c}{\multirow{7}[2]{*}{ViT-L/14}} & Pic2Word{\footnotesize (CVPR'23)} & 26.20  & 43.60  & 20.00  & 40.20  & 27.90  & 47.40  & 24.70  & 43.70  \\
    \multicolumn{2}{c}{} & SEARLE{\footnotesize (ICCV'23)} & 26.89  & 45.58  & 20.48  & 43.13  & 29.32  & 49.97  & 25.56  & 46.23  \\
    \multicolumn{2}{c}{} & SEARLE-OTI{\footnotesize (ICCV'23)} & 30.37  & 47.49  & 21.57  & 44.47  & 30.90  & 51.76  & 27.61  & 47.90  \\
    \multicolumn{2}{c}{} & LinCIR{\footnotesize (CVPR'24)} & 29.10  & 46.81  & 20.92  & 42.44  & 28.81  & 50.18  & 26.28  & 46.49  \\
    \multicolumn{2}{c}{} & CIReVL{\footnotesize (ICLR'24)} & 29.49  & 47.40  & \underline{24.79}  & 44.76  & 31.36  & \underline{53.65}  & \underline{28.55}  & 48.57  \\
    \multicolumn{2}{c}{} & LDRE{\footnotesize (SIGIR'24)} & \underline{31.04}  & \textbf{51.22}\hspace{0pt}  & 22.93  & \underline{46.76}  & \underline{31.57}  & 53.64  & 28.51  & \underline{50.54}  \\
    \multicolumn{2}{c}{} & CVSI{\footnotesize (Ours)} & \textbf{33.56}\hspace{0pt}  & \underline{50.15}  & \textbf{25.63}\hspace{0pt}  & \textbf{48.59}\hspace{0pt}  & \textbf{34.01}\hspace{0pt}  & \textbf{54.05}\hspace{0pt}  & \textbf{31.07}\hspace{0pt}  & \textbf{50.93}\hspace{0pt}  \\
    \midrule
    \multicolumn{2}{c}{\multirow{6}[2]{*}{ViT-G/14}} & Pic2Word{\footnotesize (CVPR'23)} & 33.17  & 50.39  & 25.43  & 47.65  & 35.24  & 57.62  & 31.28  & 51.89  \\
    \multicolumn{2}{c}{} & SEARLE{\footnotesize (ICCV'23)} & 36.46  & 55.35  & 28.16  & 50.32  & 39.83  & 61.45  & 34.81  & 55.71  \\
    \multicolumn{2}{c}{} & LinCIR{\footnotesize (CVPR'24)} & \textbf{46.76}\hspace{0pt}  & \textbf{65.11}\hspace{0pt}  & \underline{38.08}  & \underline{60.88}  & \textbf{50.48}\hspace{0pt}  & \textbf{71.09}\hspace{0pt}  & \textbf{45.11}\hspace{0pt}  & \underline{65.69}  \\
    \multicolumn{2}{c}{} & CIReVL{\footnotesize (ICLR'24)} & 33.71  & 51.42  & 27.07  & 49.53  & 35.80  & 56.14  & 32.19  & 52.36  \\
    \multicolumn{2}{c}{} & LDRE{\footnotesize (SIGIR'24)} & 35.94  & 58.58  & 26.11  & 51.12  & 35.42  & 56.67  & 32.49  & 55.46  \\
    \multicolumn{2}{c}{} & CVSI{\footnotesize (Ours)} & \underline{44.95}  & \underline{63.59}  & \textbf{40.75}\hspace{0pt}  & \textbf{63.71}\hspace{0pt}  & \underline{49.21}  & \underline{69.86}  & \underline{44.97}  & \textbf{65.72}\hspace{0pt}  \\
    \bottomrule
    \end{tabular}%
  \label{tab:FashionIQ}%
\end{table*}%

\subsection{Results and Analysis}
Tables~\ref{tab:CIRCO and CIRR} and~\ref{tab:FashionIQ} summarize the performance comparison on the three datasets. From these tables, we obtained the following observations: 
1) Our model achieves state-of-the-art performance across all metrics on the CIRCO and CIRR datasets and on most metrics of the FashionIQ dataset, which demonstrates the effectiveness of our model in the ZS-CIR task.
2) On CIRCO and CIRR, LLM-based methods clearly outperform textual inversion methods due to the more complex modification texts in these datasets, where the retrieval results rely more on semantic information. However, on FashionIQ, with the ViT-G/14 backbone, the textual inversion method represented by LinCIR performs significantly better than LLM-based methods. This is because the modification text in this dataset is simpler, allowing visual information to dominate the final retrieval, which limits the performance of LLM-based methods. The results align with our expectations for the real-world ZS-CIR task and our method's weight allocation strategies. 
3) Our method extracts both semantic and visual information from query and target images, and then combines these features for complementary information retrieval, which enables a balanced performance across both types of datasets (visual information-dominant and semantic information-dominant). Clearly, our method shows a significant improvement over the previous state-of-the-art method, LDRE, on the CIRCO and CIRR datasets. While the improvement on FashionIQ is limited. This is because our method extracts semantic information for both the query and the target, allowing for effective filtering of irrelevant domain pairs during retrieval. However, FashionIQ is a dataset focused on the fashion domain, which limits the effectiveness of our model on this dataset. As a result, CVSI would perform better on multi-domain datasets, aligning with the characteristics of the real-world ZS-CIR task.

\begin{figure*}[t]
  \centering
  \begin{minipage}[t]{0.3\textwidth}
    \centering
    \includegraphics[width=\linewidth]{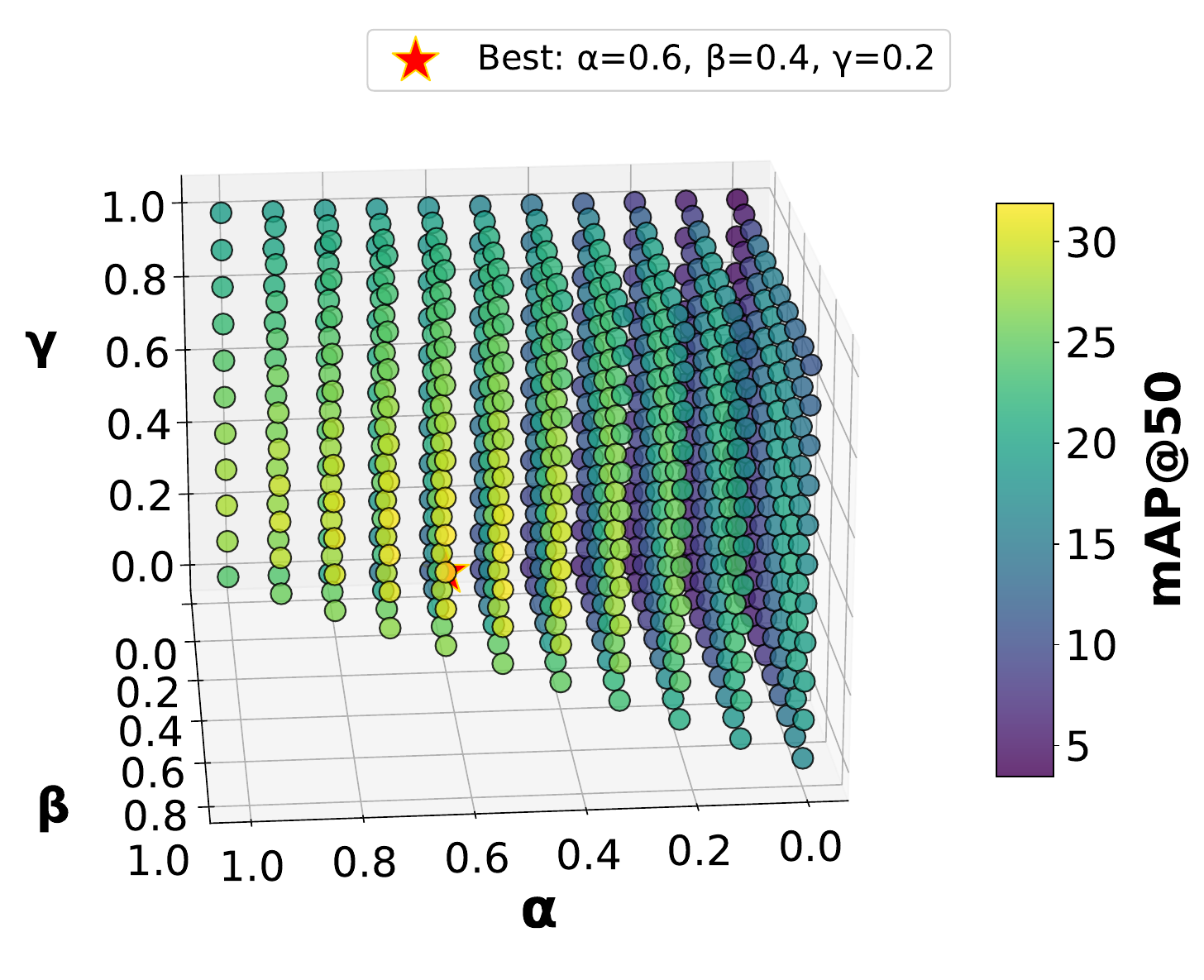}
    \\ 
    \small (a) 
  \end{minipage}
  \hfill 
  \begin{minipage}[t]{0.3\textwidth}
    \centering
    \includegraphics[width=\linewidth]{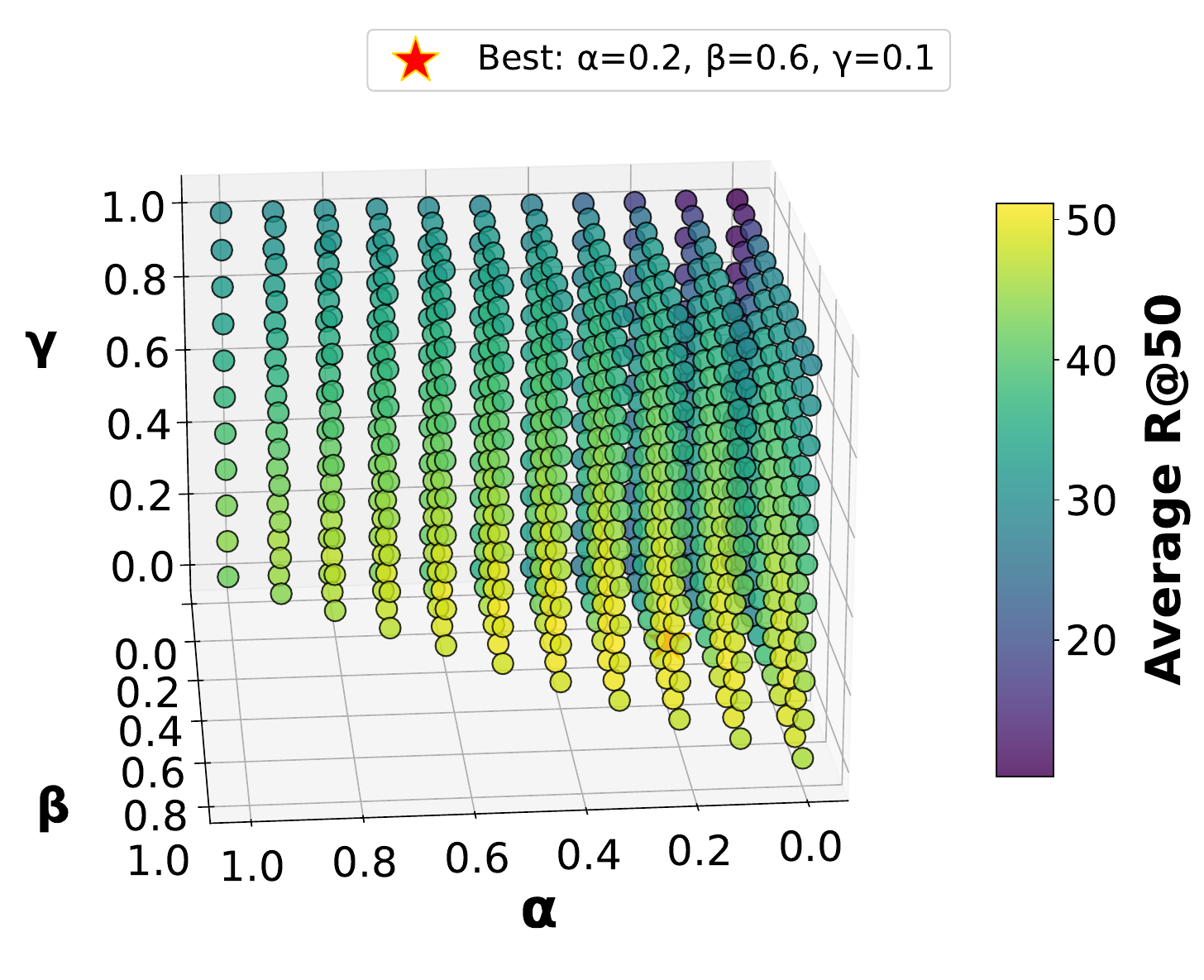}
    \\
    \small (b)
  \end{minipage}
  \hfill
  \begin{minipage}[t]{0.3\textwidth}
    \centering
    \includegraphics[width=\linewidth]{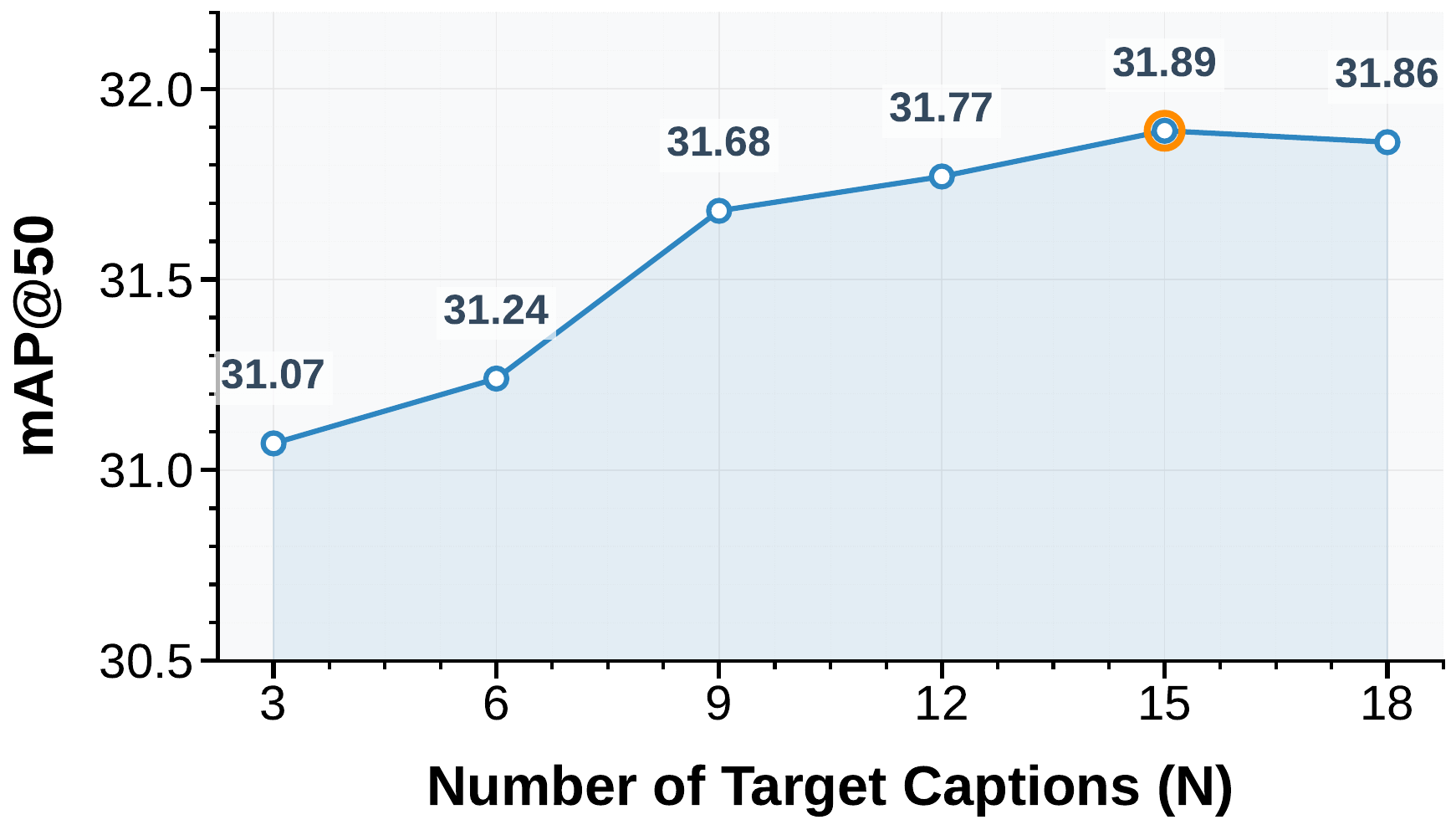}
    \\
    \small (c)
  \end{minipage}
  \caption{Impact of different hyperparameters on model performance across CIRCO and FashionIQ. (a) Model performance on CIRCO with different hyperparameter combinations. (b) Model performance on FashionIQ with different hyperparameter combinations. (c) Model performance on CIRCO with different numbers of target captions.} 
  \label{figure4}
\end{figure*}

\subsection{Ablation Studies}
To evaluate the contribution of each component of our model and the effectiveness of the features involved in complementary information retrieval, we conducted ablation experiments. We compared CVSI with the following derivatives: 1) \textit{w/o visual information extraction}: The query and target do not undergo visual information extraction, and only semantic information is involved in complementary information retrieval. 2) \textit{w/o semantic information extraction}: The query and target do not undergo semantic information extraction, and only visual information is involved in complementary information retrieval. 3) \textit{w/o reference image features}: The reference image features are excluded from the complementary information retrieval process. 4) \textit{w/o fine-grained features}: Fine-grained features are excluded from the complementary information retrieval process. 5) \textit{w/o average modified captions features}: The average modified captions features of the query are excluded from the complementary information retrieval process. 6) \textit{w/o target image features}: The target image features are excluded from the complementary information retrieval process. 7) \textit{w/o target average captions features}: The average captions features of the target are excluded from the complementary information retrieval process.

\begin{table}[t]
  \centering
  \caption{Ablation studies on CIRCO test set and FashionIQ validation set with ViT-L/14 backbone.}
  \small
    \setlength{\tabcolsep}{4.5pt} 
    \begin{tabular}{lll|c|c}
    \toprule
    \multicolumn{3}{c|}{\multirow{3}{*}{Method}} 
    & \multirow{2}{*}{\makecell{CIRCO}} 
    & \multirow{2}{*}{\makecell{FashionIQ\\ (Avg)}} \\
    \multicolumn{3}{c|}{} & & \\ 
    \cmidrule{4-5}
    \multicolumn{3}{c|}{} & mAP@10 & R@10 \\
    \midrule
    \multicolumn{3}{l|}{w/o visual information extraction} & 16.75  & 15.81  \\
    \multicolumn{3}{l|}{w/o semantic information extraction} & 14.15  & 28.44  \\
    \multicolumn{3}{l|}{w/o reference image features} & 28.24  & 28.11  \\
    \multicolumn{3}{l|}{w/o fine-grained features} & 24.96  & 20.63  \\
    \multicolumn{3}{l|}{w/o average modified captions features} & 17.16  & 29.39  \\
    \multicolumn{3}{l|}{w/o target images features } & 19.67  & 16.08  \\
    \multicolumn{3}{l|}{w/o target average captions features} & 23.24  & 29.54  \\
    \multicolumn{3}{l|}{\textbf{CVSI}} & \textbf{28.24} & \textbf{31.07} \\
    \bottomrule
    \end{tabular}%
  \label{tab:Ablation}%
\end{table}%

Table \ref{tab:Ablation} presents the results of the ablation experiments on CIRCO and FashionIQ datasets. From this table, we can make the following observations: 1) When CVSI lacks a component or a feature in complementary information retrieval, the model's performance deteriorates, which highlights the effective role of each component in CVSI and the importance of the features involved in the final retrieval process. 2) On CIRCO dataset, the worst results occur when semantic information is excluded from the retrieval process, while on FashionIQ dataset, the worst results occur when visual information is excluded. It illustrates that the roles of visual and semantic information differ depending on the nature of the dataset, confirming the necessity of weighting these features in CVSI. 3) An interesting case is the reference image features. Excluding these features from the final retrieval decreases performance on FashionIQ, but does not affect CIRCO. It is because, for CIRCO dataset, we set the reference image features weight to 0, as shown in the subsequent hyperparameter analysis experiments. 4) When the target image's features are not involved in the final retrieval, there is a significant drop in the model’s performance. Similarly, the absence of target average caption features also leads to a decline in performance. This indicates that extracting information from both aspects of the target is essential, and the textual information in the target can assist the visual information. In conclusion, the components and features of CVSI have a significant impact on the model's final performance, with the extent of their influence varying depending on the nature of the dataset (e.g., whether semantic or visual information predominates).

\begin{table}[t]
  \centering
  \caption{Experimental results of different LLMs on CIRCO mAP@k metric.}
    \begin{tabular}{c|cc|cccc}
    \toprule
    Arch  & \multicolumn{2}{c|}{LLM} & k=5   & k=10  & k=25  & k=50 \\
    \midrule
    \multirow{3}[2]{*}{ViT-B/32} & \multicolumn{2}{c|}{LLama2-70B} & 20.87  & 21.58  & 23.52  & 24.46  \\
          & \multicolumn{2}{c|}{GPT-3.5-Turbo} & 21.67  & 22.47  & 24.46  & 25.44  \\
          & \multicolumn{2}{c|}{GPT-4o} & 22.45  & 23.20  & 25.28  & 26.29  \\
    \midrule
    \multirow{3}[2]{*}{ViT-L/14} & \multicolumn{2}{c|}{LLama2-70B} & 25.84  & 26.99  & 29.52  & 30.56  \\
          & \multicolumn{2}{c|}{GPT-3.5-Turbo} & 26.89  & 28.24  & 30.79  & 31.89  \\
          & \multicolumn{2}{c|}{GPT-4o} & 27.63  & 28.91  & 31.59  & 32.66  \\
    \midrule
    \multirow{3}[2]{*}{ViT-G/14} & \multicolumn{2}{c|}{LLama2-70B} & 32.50  & 33.89  & 36.78  & 37.89  \\
          & \multicolumn{2}{c|}{GPT-3.5-Turbo} & 33.51  & 34.96  & 37.86  & 38.98  \\
          & \multicolumn{2}{c|}{GPT-4o} & 35.03  & 36.48  & 39.55  & 40.68  \\
    \bottomrule
    \end{tabular}%
  \label{tab:LLMs}%
\end{table}%

\begin{table*}[t]
  \centering
  \caption{Experimental Comparison of Method Efficiency and Performance Based on the ViT-L/14 Backbone}
    \begin{tabular}{lccccccccccc}
    \toprule
    \textbf{Method} & \multicolumn{2}{c}{\textbf{Training time(h)}} & \multicolumn{3}{c}{\textbf{BLIP2+LLM inference time(s)}} & \multicolumn{2}{c}{\textbf{Retrieval time(s)}} & \multicolumn{2}{c}{\textbf{CIRCO(mAP@10)}} & \multicolumn{2}{c}{\textbf{CIRR(Recall@10)}} \\
    \midrule
    Pic2Word & \multicolumn{2}{c}{$>$5} & \multicolumn{3}{c}{——} & \multicolumn{2}{c}{0.02} & \multicolumn{2}{c}{9.51} & \multicolumn{2}{c}{65.3} \\
    SEARLE & \multicolumn{2}{c}{$>$5} & \multicolumn{3}{c}{——} & \multicolumn{2}{c}{0.02} & \multicolumn{2}{c}{12.73} & \multicolumn{2}{c}{66.29} \\
    LinCIR & \multicolumn{2}{c}{1.5} & \multicolumn{3}{c}{——} & \multicolumn{2}{c}{0.02} & \multicolumn{2}{c}{13.58} & \multicolumn{2}{c}{66.68} \\
    CIReVL & \multicolumn{2}{c}{——} & \multicolumn{3}{c}{1.6} & \multicolumn{2}{c}{0.03} & \multicolumn{2}{c}{19.01} & \multicolumn{2}{c}{64.92} \\
    LDRE  & \multicolumn{2}{c}{——} & \multicolumn{3}{c}{23.9} & \multicolumn{2}{c}{0.05} & \multicolumn{2}{c}{24.03} & \multicolumn{2}{c}{67.54} \\
    \textbf{CVSI}  & \multicolumn{2}{c}{1.5} & \multicolumn{3}{c}{24.7} & \multicolumn{2}{c}{0.05} & \multicolumn{2}{c}{\textbf{28.24}} & \multicolumn{2}{c}{\textbf{75.04}} \\
    \bottomrule
    \end{tabular}%
  \label{tab:Efficiency}%
\end{table*}%

\subsection{Hyperparameter Analysis}
In this section, we analyze the impact of different hyperparameters on model performance using a ViT-L/14 backbone. As shown in Figure \ref{figure4}, we observe that the choice of hyperparameters significantly influences the model's performance, and our key findings are as follows: 1) As shown in Figure \ref{figure4}(a), for datasets like CIRCO, which are dominated by semantic information, the optimal hyperparameter combination is $\alpha=0.6$, $\beta=0.4$, $\gamma=0.2$. When the weight of semantic information ($\alpha$) is higher, the model performs better. 2) As shown in Figure \ref{figure4}(b), for datasets like FashionIQ, which are dominated by visual information, the optimal hyperparameter combination is $\alpha=0.2$, $\beta=0.6$, $\gamma=0.1$. It is evident that when the weight of visual information ($\beta$) is larger, the model achieves better performance. 3) The value of $\gamma$ is generally low, which can be attributed to CLIP being pre-trained on text-image pairs, where its ability to contrast semantics between texts is relatively weak. As a result, the target caption features serve mainly as auxiliary information. 4)  For the number of captions generated for the query, we selected the value of $N$ as in \cite{ref27}. in Figure \ref{figure4}(c), for the number of captions generated for the target, we find that when $N=15$, the model's performance also reaches its peak. The balanced value enriches the semantic information of the target images while avoiding semantic divergence caused by an excessive number of captions.

\subsection{Scalability Experiments of LLMs}
Considering that our model relies on LLMs to generate modified captions, and the quality of these captions directly affects the final retrieval results, we conducted scalability experiments on the selection of LLMs. The results are presented in table \ref{tab:LLMs}. We observe that the choice of LLM significantly impacts the performance, with the model using the open-source LLama2-70B performing worse than the closed-source GPT models. Among these, GPT-4o, with its superior generation capabilities, yields the best performance. Specifically, when using the ViT-L/14 backbone, GPT-4o results in an average improvement of 2.53\% across the top-k metrics compared to GPT-3.5-Turbo (e.g., from 26.89 to 27.63 at k=5, and from 31.89 to 32.66 at k=50). Similarly, when using the ViT-G/14 backbone, GPT-4o outperforms GPT-3.5-Turbo by an average of 4.43\%. This trend suggests that as the capabilities of the LLM improve, our model's performance also improves, with GPT-4o consistently delivering superior results across both backbones.

\subsection{Efficiency and Performance Comparison}
In practical retrieval applications, it is necessary to explore the trade-off between retrieval efficiency and performance. Therefore, we performed experiments to compare CVSI with previous methods in terms of efficiency and retrieval performance using the same experimental setup. Training time refers to the total time for model training, while inference time and retrieval time refer to the time taken for a single retrieval query. The experimental results are shown in Table \ref{tab:Efficiency}. We can observe that among the two categories of methods, although LLM-based methods show reduced efficiency compared to textual inversion methods, they achieve a significant improvement in retrieval performance. This trade-off is meaningful: users typically prefer to find the correct target image in a single, slightly longer but successful query, rather than conducting multiple fast yet ineffective attempts. Moreover, we see that while CVSI introduces a slight increase in inference time compared to the prior sota method LDRE (+0.8s), this small cost results in a significant performance boost: CIRCO mAP@10 increases by 17.52\% and CIRR Recall@10 improves by 11.10\%, which shows that CVSI provides a high return on performance improvement. Additionally, model training and target feature generation can be done before actual retrieval, minimizing the impact on real-world retrieval tasks. This further demonstrates CVSI's good balance between efficiency and performance.

\begin{figure}[t]
  \centering
  \includegraphics[width=\linewidth]{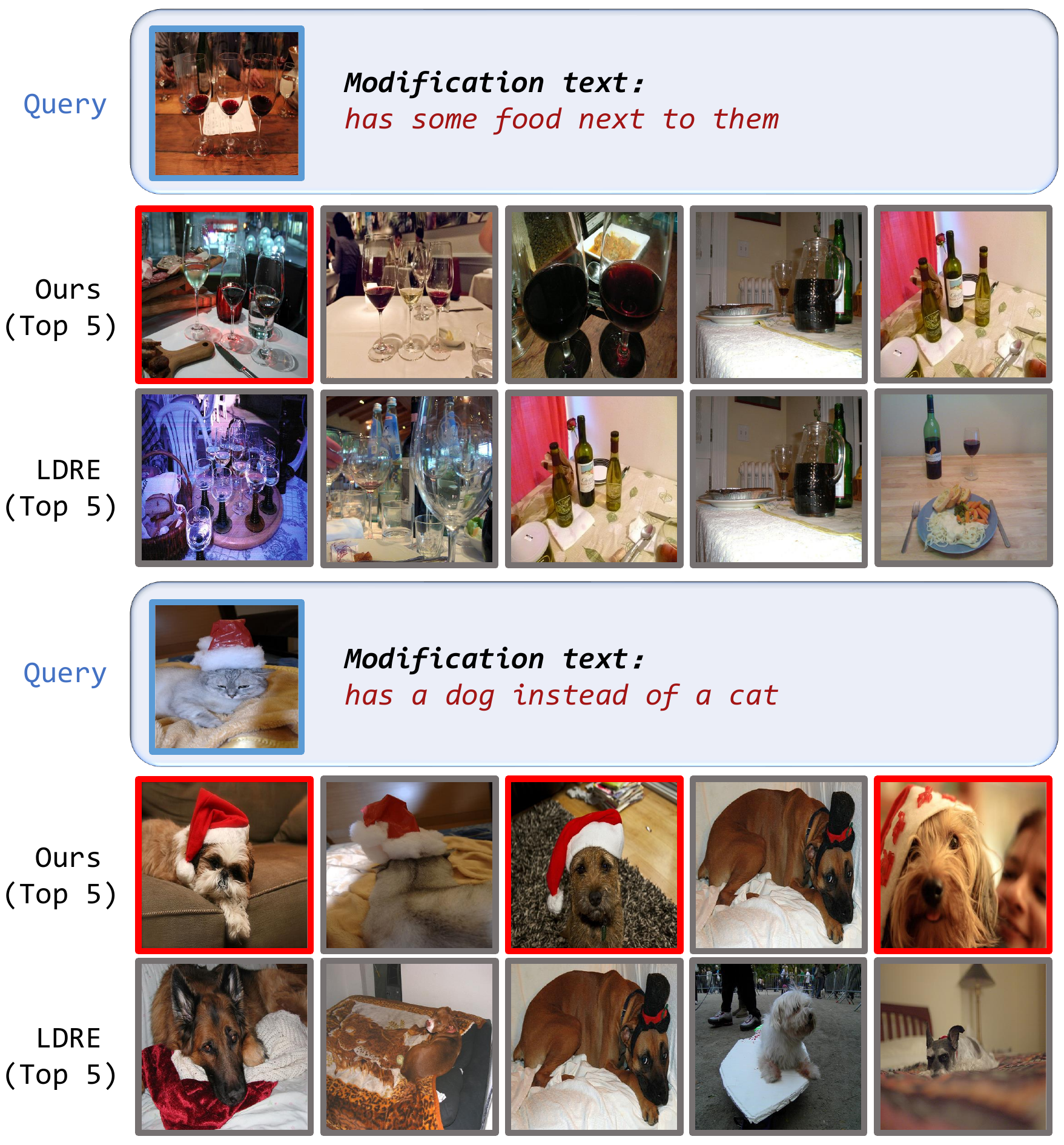}
  \caption{Qualitative analysis of top-5 retrieval results by our method and LDRE on the CIRCO test set, with ground-truth images highlighted in red.}
  \label{figure 5}
\end{figure}

\subsection{Qualitative Results}
Figure \ref{figure 5} presents a qualitative comparison between our method and LDRE (previous sota method). We select several examples from the CIRCO test set and visualize the top-5 retrieval results. We observe that our method effectively achieves a complementary integration of visual and semantic information. As shown in Figure \ref{figure 5}, in the upper example, LDRE fails to accurately identify the shape and number of the wine glasses. In the lower example, it also fails to correctly recognize the Santa hat, which is an essential element that should have been retained. This is primarily because LDRE relies solely on captions to convert visual information. However, language often struggles to intuitively convey the full appearance of an object and cannot fully describe all the details present in a reference image. As a result, the visual representation generated in this way is inherently incomplete. In contrast, our method is capable of effectively integrating comprehensive visual and semantic information, and it can extract fine-grained object-level changes. The vertical comparison of results demonstrates that such complete extraction and effective integration enable our approach to achieve significantly more accurate retrieval performance.

\section{Conclusion}
In this work, we propose CVSI, a fine-grained method that integrates both visual and semantic information to address ZS-CIR task. Our approach overcomes the limitations of previous methods used for ZS-CIR, by designing a visual information extraction component to capture more comprehensive visual features and a semantic information extraction component to gather rich semantic insights. Furthermore, by focusing on specific object changes, we enhance the fine-grained retrieval capability. Ultimately, we extract information from both the query and the target, matching them through complementary information retrieval, allowing the model to handle the complex real-world CIR task. Extensive experiments conducted on three benchmark datasets demonstrate that our model achieves new state-of-the-art results across most metrics. In the future, we plan to explore further applications in the fields of multimodal retrieval and fusion. 

\section*{Acknowledgment}
This research was partially supported by the National Natural Science Foundation of China (U23A20319, 62441239,2406303), Anhui Provincial Natural Science Foundation (No. 2308085QF229), Anhui Province Science and Technology Innovation Project (202423k09020010), the Fundamental Research Funds for the Central Universities (No. WK2150110034).

\bibliographystyle{IEEEtran}
\bibliography{IEEEabrv,references.bib}

\end{document}